\documentclass{article}

\usepackage[preprint]{nips_2018}

\usepackage[utf8]{inputenc} % allow utf-8 input
\usepackage[T1]{fontenc}    % use 8-bit T1 fonts
\usepackage{hyperref}       % hyperlinks
\usepackage{url}            % simple URL typesetting
\usepackage{booktabs}       % professional-quality tables
\usepackage{amsfonts}       % blackboard math symbols
\usepackage{nicefrac}       % compact symbols for 1/2, etc.
\usepackage{microtype}      % microtypography
\usepackage{amsmath}
\usepackage{graphicx}
\usepackage{algorithm}% http://ctan.org/pkg/algorithms
\usepackage{algpseudocode}% http://ctan.org/pkg/algorithmicx
\usepackage{wrapfig}
\usepackage{multirow}
\usepackage[T1]{fontenc}
\usepackage{tikz}
\usetikzlibrary{arrows,shapes,positioning,shadows,trees}
\usepackage{xcolor}
\usepackage{dashbox}

\newcommand{\eval}{\textsc{eval}}
\newcommand{\evalo}{\textbf{\texttt{evalo}}}
\newcommand{\lookupo}{\textbf{\texttt{lookupo}}}

\newcommand{\car}{\texttt{car}}
\newcommand{\cdr}{\texttt{cdr}}
\newcommand{\cons}{\texttt{cons}}

\newcommand{\lvar}[1]{%
    \colorlet{currentcolor}{.}%
    {\color{blue}%
    \setlength{\fboxsep}{1pt}\dbox{\color{darkgray}\texttt{#1}}}%
}
\newcommand{\expr}{\texttt{P}}
\newcommand{\env}{\texttt{I}}
\newcommand{\out}{\texttt{O}}

\newcommand{\disj}{\textsc{disj}}
\newcommand{\arr}{\;\rightarrow}
\usepackage[title]{appendix}

\title{Neural Guided Constraint Logic Programming for Program Synthesis}

\author{Lisa Zhang\textsuperscript{1,2},
        Gregory Rosenblatt\textsuperscript{4},
        Ethan Fetaya\textsuperscript{1,2},
        Renjie Liao\textsuperscript{1,2,3},
        William E. Byrd\textsuperscript{4}, \\
\textbf{Matthew Might\textsuperscript{4},
        Raquel Urtasun\textsuperscript{1,2,3},
        Richard Zemel\textsuperscript{1,2}} \\
\textsuperscript{1}University of Toronto, \textsuperscript{2}Vector Institute, \textsuperscript{3}Uber ATG, \textsuperscript{4}University of Alabama at Birmingham\\
\textsuperscript{1}\texttt{\{lczhang,ethanf,rjliao,urtasun,zemel\}@cs.toronto.edu} \\
\textsuperscript{4}\texttt{\{gregr,webyrd,might\}@uab.edu}
}

\begin{document}

\maketitle

\begin{abstract}
    Synthesizing programs using example input/outputs is a classic problem in artificial intelligence.
    We present a method for solving Programming By Example (PBE) problems by
    using a neural model to guide the search of a constraint logic programming system
    called miniKanren. Crucially, the neural model uses miniKanren's internal
    representation as input; miniKanren represents a PBE problem as recursive
    constraints imposed by the provided examples.
    We explore Recurrent Neural Network and Graph Neural Network models.
    % which use these constraints as input to score candidate programs.
    We contribute a modified miniKanren, drivable by an external agent,
    available at \href{https://github.com/xuexue/neuralkanren}{https://github.com/xuexue/neuralkanren}.
    We show that our neural-guided approach using constraints can
    synthesize programs faster in many cases, and importantly, can
    generalize to larger problems.
\end{abstract}

\section{Introduction}

Program synthesis is a classic area of artificial intelligence
that has captured the imagination of many computer scientists.
Programming by Example (PBE) is one way to formulate
program synthesis problems, where example input/output pairs 
specify a target program.
In a sense, supervised learning can be considered program
synthesis, but supervised learning via successful models like
deep neural networks famously lacks interpretability.
The clear interpretability of programs as code
means that synthesized results can be compared,
optimized, translated, and proved correct.
The manipulability of code makes
program synthesis continue to be relevant today.

Current state-of-the-art approaches use symbolic techniques developed by the
programming languages community. These
methods use rule-based, exhaustive search, often
manually optimized by human experts. % using domain-specific knowledge.
While these techniques excel for small problems, they tend not to scale.
% While these techniques work, they tend not to scale, and success is limited to small programs.
Recent works by the machine learning community explore a
variety of statistical methods to solve PBE problems more quickly.
Works generally fall under three categories:
differentiable programming \cite{neelakantan2015neural,reed2015neural,graves2014neural}, 
direct synthesis \cite{devlin2017robustfill,parisotto2016neuro},
and neural guided search \cite{balog2016deepcoder,kalyan2018neuralguided}.

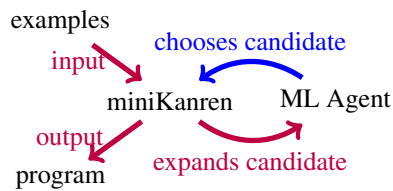
\begin{wrapfigure}{r}{7cm}
    \vspace{-20pt}
    \centering
    \begin{tikzpicture}[transform shape,line width=2pt]
      \tikzset{every node/.style={node distance=2.2cm}}
      \node (in) {examples};
      \node (mk) [below right of = in, yshift=0.5cm,xshift=-0.1cm] {miniKanren};
      \node (ml) [right of = mk] {ML Agent} ;
      \node (out) [below left of= mk, yshift=0.5cm,xshift=0.1cm] {program};
      \path (mk) edge [->, purple, bend right=35] node [below] {expands candidate} (ml)
            (ml) edge [->, blue, bend right=35] node [above] {chooses candidate} (mk)
            (in) edge [->, purple] node [left] {input} (mk)
            (mk) edge [->, purple] node [left] {output} (out)
        ;
    \end{tikzpicture}
    \caption{Neural Guided Synthesis Approach}
    \label{fig:setup}
    \vspace{-6pt}
\end{wrapfigure}

This work falls under neural guided search, where the machine learning
model guides a symbolic search.
We take integrating with a symbolic system further: we use
its internal representation as input to the neural model.
%the internal representation of the symbolic system as input to the neural model.
The symbolic system we use is a constraint logic programming system
called miniKanren\footnote{%
The name ``Kanren'' comes from the Japanese word for ``relation''.
}%
\cite{byrd2006}, chosen for its ability to encode synthesis
problems that are difficult to express in other systems.
Specifically, miniKanren does not rely on types, is able to
to complete partially specified programs, and has a
straightforward implementation~\cite{hemann2013mukanren}.
%flexibility ==> dynamic typing, ability to fill in holes
%chosen for its potential to synthesize
%We choose miniKanren for its potential to synthesize
%recursive programs in dynamically typed languages .
%This flexibility allows miniKanren 
miniKanren searches for a candidate program that satisfies the recursive
constraints imposed by the input/output examples.
Our model uses these constraints to score candidate programs and guide
miniKanren's search.

%Second, its internal
%representation of the problem is a tree structure that encodes
%not only the search space, but also the input/output pairs. This
%allows us to not have an encoder for input/output pairs, and work
%directly on the search tree itself using a
%Gated Graph Neural Network (GGNN) \cite{li2015gated} search
%strategy.

%This idea of using the constraint tree representation is
%equivalent to learning the specific flavour of constraint
%satisfaction miniKanren uses for program synthesis.
%That is, the goal of the neural model is to learn
%which candiate partial program
%has the most promise, or the most probability of fully satisfying
%the constraints imposed by input/output examples.
%The symbolic solver follows the output of the neural model to
%explore the search space.

Neural guided search using constraints
is promising for several reasons. First, while symbolic approaches
outperform statistical methods, they have not demonstrated an ability to
scale to larger problems; neural guidance may %offers the potential to
help navigate the exponentially growing search space.
%First, symbolic systems have
%historically outperformed purely statistical approaches, though neural guidance
%can help navigate exponentially large search spaces.
Second, symbolic systems exploit the compositionality of synthesis
problems: miniKanren's constraints select portions of the input/output
examples relevant to a subproblem, akin to having a symbolic attention
mechanism. Third, constraint lengths are relatively stable even as
we synthesize more complex programs; our approach
should be able to generalize to programs larger than those seen in training.
%Guiding a search and using constraints both alleviate this problem.
%We present some evidence that our approach is able to generalize to programs
%larger than those seen in training.

To summarize, we contribute a novel form of neural guided synthesis, where we 
use a symbolic system's internal representations to solve an auxiliary
problem of constraint scoring using neural embeddings.
We explore two models for scoring constraints: 
Recurrent Neural Network (RNN) and Graph Neural Network (GNN)
\cite{scarselli2009graph}.
%The use of GNN to represent syntactical structures also appeared in concurrent
%works like \cite{allamanis2018learning}, which used graph neural
%networks to represent programs, and \cite{selsam2018learning},
%which used graph neural networks to solve boolean satisfiability (SAT)
%problems. Though our ideas are similar to those two works, the details
%of the implementation are different.
We also present a ``transparent'' version of miniKanren with
visibility into its internal constraints, available at
\href{https://github.com/xuexue/neuralkanren}{https://github.com/xuexue/neuralkanren}.

%With these new ideas, we show
%evidence that our method has the potential to generalize to larger problems.
%We also contribute a version of miniKanren that provides visibility into its
%internal representation, and allows an external agent to guide search.

Our experiments focus on synthesizing programs in a subset of Lisp,
and show that scoring constraints help.
%We test our approach to solve PBE problems in a subset of Lisp,
%and show improved performance on generated problems held out from training.
%We compare three variations of neural-guided models
%against two unguided miniKanren search strategies on problems held out from training.
More importantly, we test the generalizability of our approach on three families
%To gauge generalizability, we test our approaches on three families
of synthesis problems. We compare against
state-of-the-art systems $\lambda^2$~\cite{feser2015synthesizing},
Escher~\cite{albarghouthi2013recursive}, Myth~\cite{osera2015type}, and
RobustFill~\cite{devlin2017robustfill}.
We show that our %neural-guided approach using constraints can
%synthesize problems faster in many cases, and
approach has the potential to generalize
to larger problems.

%First, on list manipulation problems in a subset of the Lisp programming
%language. Second, we solve a subset of FlashFill like string
%maniuplation problems used in the SyGus competition 2017 \cite{alur2017sygus}.
%Lastly, to demonstrate generality, we solve recursive problems
%in Lisp.

\section{Related Work}

% While there is a recent surge in interest in program synthesis,
Programming by example (PBE) problems have a long history dating
to the 1970's \cite{summers1977methodology, biermann1978inference}.
Along the lines of early works in program synthesis, the
programming languages community developed search
techniques that enumerate possible programs, with pruning strategies
based on types, consistency, and logical reasoning to improve the search.
%While such techniques
%may seem simple, they can achieve impressive results.
Several state-of-the-art methods are described in Table~\ref{tab:rel}.

\begin{table}[h]
\centering
\caption{Symbolic Methods}
\label{tab:rel}
\begin{tabular}[h]{llll}%l}
\hline
Method & Direction & Search Strategy & Type Discipline \\ % & Focus \\
\hline
miniKanren \citep{byrd2006, byrd2017} % \citep{byrd2012miniKanren} % 
       & Top-down         & Biased-Interleaving
                                            & Dynamic  \\ % & Symbols  \\
$\lambda^2$ \citep{feser2015synthesizing}
       & Top-down         & Template Complexity
                                            & Static   \\ %& Numeric Lists \\
Escher \citep{albarghouthi2013recursive}
       & Bottom-up        & Forward Search / Conditional Inference
                                            & Static   \\ %& Active Learning \\
Myth \citep{osera2015type}
       & Top-down         & Iterative Deepening
                                            & Static   \\%& Parsiomny \\
\hline
\end{tabular}
\end{table}

The method $\lambda^2$ \cite{feser2015synthesizing} is most similar to
miniKanren, but specializes in numeric, statically-typed inputs and outputs.
Escher \cite{albarghouthi2013recursive}
is built as an active learner, and relies on the presence of an oracle
to supply outputs for new inputs that it chooses.
Myth \cite{osera2015type} searches for the smallest program satisfying a set
of examples, and guarantees parsimony.
These methods all use functional languages based on the $\lambda$-calculus
as their target language, and aim to
synthesize general, recursive functions.

Contributions by the machine learning community have grown in the
last few years. Interestingly, while PBE problems can be thought of
as a meta-learning problem, few works
explore this relationship. Each synthesis problem can be thought
of as a learning problem~\cite{chen2018towards}, so learning the synthesizer
can be thought of as meta-learning.
Instead, works generally fall under direct synthesis, differentiable programming,
and neural guided synthesis.
%This section summarizes the recent contributions in these areas.

\paragraph{Direct Synthesis} 
In direct synthesis, the program is produced directly as a sequence or tree.
One domain where this has been successful is
string manipulation as applied to spreadsheet completion,
as in FlashFill \cite{gulwani2011flashfill} and its descendants
\cite{parisotto2016neuro, devlin2017robustfill, bhupatiraju2017deep}.
%These techniques %generates a program as a sequence or tree,
%conditioned on the input/output examples. They
%are trained on string manipulation tasks and are
%successfully applied to spreadsheet completion problems.
FlashFill \cite{gulwani2011flashfill} 
uses a combination of search and carefully crafted heuristics.
Later works like \cite{parisotto2016neuro} introduce
a ``Recursive-Reverse-Recursive Neural Network''
to generate a program tree conditioned on input/output embeddings.
More recently, RobustFill \cite{devlin2017robustfill} uses
bi-directional Long Short-Term Memory (LSTM) with attention, to generate programs as
sequences. Despite flattening the tree structure,
RobustFill achieved much better results (92\% vs 38\%) on the FlashFill benchmark.
While these approaches succeed in the practical domain of string manipulation,
we are interested in exploring manipulations of richer data structures.

%While practical, the approach to string manipulation is not a stepping
%stone to general computation, and we wish to work with more nuanced
%data structures.
%is greatly limited, so methods developed
%for this domain may not scale to the larger problem of synthesizing
%programs involving recursion.

\paragraph{Differentiable Programming}
Differentiable programming involves building a differentiable interpreter,
then backpropagating through the interpreter to learn a \textit{latent}
program. The goal is to infer correct outputs for new inputs.
Work in differentiable programming began with the Neural Turing
Machine \cite{graves2014neural}, a neural architecture that augments
neural networks with external memory and attention.
%emulating a
%von Neumann architecture in an end-to-end differentiable manner.
Neural Programmer \cite{neelakantan2015neural} and 
Neural Programmer-Interpreter \cite{reed2015neural} extend the work
with reusable operations, and build programs compositionally.
%Neural Programmer \cite{neelakantan2015neural} builds a similar
%neural architecture augmented using basic operations, creating
%complex programs compositionaly.
%The Neural Programmer-Interpreter \cite{reed2015neural} is also
%compositional, with %three separate learnable components that incldue
%a recurrent core, a key-value memory of program embeddings,
%and a task-dependent encoder. % It is trained on full program traces.
%The RobustFill work \cite{devlin2017robustfill} also explores
%using a latent program instead of an explicit one. They find that
%latent programs is more likely to get some inputs correct, but explicit
%programs generalize better when successful.
While differentiable approaches are appealing,
\cite{gaunt2016terpret} showed that this approach
still underperforms discrete search-based techniques.

\paragraph{Neural Guided Search}
A recent line of work uses statistical techniques to guide a
discrete search.
For example, DeepCoder \cite{balog2016deepcoder} 
uses an encoding of the input/output examples to
predict functions that are likely to appear in the program, to
prioritize programs containing those functions.
More recently, \cite{kalyan2018neuralguided} uses an LSTM to guide
the symbolic search system PROSE
(Microsoft Program Synthesis using Examples).
The search uses a ``branch and bound'' technique. The neural model
learns the choices that maximize the bounding function $h$
introduced in \cite{gulwani2011flashfill} and used
for FlashFill problems.
These approaches attempt to be search system agnostic, whereas
we integrate deeply with one symbolic approach,
taking advantage of its internal representation
and compositional reasoning.

Other work in related domains shares similarities with our contribution.
For example, \cite{ellis2016sampling} uses constraint-based solver
to sample terms in order to complete a program sketch, but is not concerned with
synthesizing entire programs.
Further, \cite{rocktaschel2017end} implements differentiable logic programming
to do fuzzy reasoning and induce soft inference rules.
They use Prolog's depth-first search as-is and learn constraint
validation (approximate unification), whereas we learn the search strategy
and use miniKanren's constraint validation as-is.

\section{Constraint Logic Programming with miniKanren}

This section describes the constraint logic programming language
miniKanren and its use for program synthesis.
Figure~\ref{fig:setup} summarizes the relationship between miniKanren
and the neural agent.

%We also present
%the transparent implementation of miniKanren suitable for use
%by researchers working on other neural guided search models.

\subsection{Background}

The constraint logic programming language miniKanren uses the relational
programming paradigm, where programmers write \textit{relations}
instead of functions.
% A relation defines a set of ordered tuples.
%For example, the greater than relation $>$ is a binary relation,
%and all tuples $(a, b)$ where $a > b$ belong to this relation.
Relations are a generalization of functions: a function $f$ with $n$
parameters can be expressed as a relation $R$ with $n+1$ parameters,
e.g., $(f\ x) = y$ implies $(R\ x\ y)$.
The notation  $(R\ x\ y)$ means that $x$ and $y$
are related by $R$. %Other examples of relations include $>$, $<$, and the equality relation $==$.

%miniKanren supports queries where any input to a relation can be unknown.
In miniKanren queries, data flow is not directionally biased: any input to a relation can be unknown.
%For example, a query for $\lvar{X} > 0$, where \lvar{X} is an unknown or a
%\textit{logic variable}, uses the definition of the relation $>$
%to find all corresponding values of \lvar{X}
%where \lvar{X}, and $0$  are related by $>$.
For example, % given $R$ and $f$ defined as before,
a query $(R\ \lvar{X}\ y)$ where $y$ is known and \lvar{X} is an unknown, called a \textit{logic variable},
finds values \texttt{X} where \texttt{X} and $y$ are related by $R$. In other words, given $R$ and $f$
defined as before, the query finds inputs \texttt{X} to $f$ such that $(f\ \texttt{X}) = y$.
This property allows the relational translation of a function to run
computations in reverse~\cite{byrd2017}.
We refer to such uses of relations containing logic variables as \textit{constraints}.

\begin{wrapfigure}{r}{7cm}
 \vspace{-12pt}
 \centering
     \begin{tabular}{p{0.3cm} p{0.3cm} p{0.3cm} l}
      $\texttt{(evalo \lvar{P} I O)}$ \\
      $\ \ \ \ \ \Rightarrow$ &
          \disj & $\arr$ & $\texttt{(evalo (quote \lvar{A}) I O)}$ \\
      & & $\arr$ & $\texttt{(evalo (car \lvar{B}) I O)}$ \\
      & & $\arr$ & $\texttt{(evalo (cdr \lvar{C}) I O)}$ \\
      & & $\arr$ & $\texttt{(evalo (cons \lvar{D} \lvar{E}) I O)}$ \\
      & & $\arr$ & $\texttt{(evalo (var \lvar{F}) I O)}$ \\
      & & & \dots \\
     \end{tabular}

 \caption{Expansion of an \evalo\ constraint}
 \label{fig:expand}

 % TODO: the \p seems to be breaking this caption.
 %\caption{Expansions and reduction of $(\evalo\ \lvar{\p}\ (\ii)\ \oo)$}
\end{wrapfigure}

In this work, we are interested in using a relational form \evalo\ of
an interpreter \eval\ to perform program synthesis\footnote{%
In miniKanren convention, a relation is named after the corresponding function, with an `o' at the end.
Appendix~\ref{sec:reln} provides a definition of \evalo\ used in our experiments.
}.
%$(\eval\ \expr\ \env) = \out$ expressed as a relation $(\evalo\ \expr\ \env\ \out)$
%We will have known inputs and outputs, and have an unknown
%\lvar{\expr} that we wish to find.
%\subsection{Expressing PBE problems}
In the functional computation $(\eval\ \expr\ \env) = \out$, program \expr\ and input \env\
are known, and the output \out\ is the result to be computed.
The same computation can be expressed relationally with $(\evalo\ \expr\ \env\ \lvar{\out})$
where \expr\ and \env\ are known and \lvar{\out} is an unknown.
%Aside from computing outputs from programs and inputs,
We can also synthesize programs from inputs and outputs,
expressed relationally with $(\evalo\ \lvar{\expr}\ \env\ \out)$ where \lvar\expr\ is unknown while \env\ and \out\ are known.
While ordinary evaluation is deterministic, there may be many valid programs \expr\ for any pair of \env\ and \out.
Multiple uses of \evalo, involving the same \lvar{\expr} but different pairs \env\ and \out\,
can be combined in a conjunction, further constraining \lvar{\expr}. %the space of valid programs to find those that generalize better. 
This is how PBE tasks are encoded using an implementation of \evalo\ for the target
synthesis language.

A miniKanren program internally represents a query as a constraint tree built out of conjunctions,
disjunctions, %equality constraints,
and calls to relations (constraints).
A relation like \evalo\ is recursive, that is,
defined in terms of invocations of other constraints including itself.
Search involves unfolding a recursive constraint
by replacing the constraint with its definition in terms of other constraints.
For example, in a Lisp interpreter, a program \expr\
can be a constant, a function call, or another expression.
Unfolding reveals these possibilities as clauses of a disjunction that replaces \evalo.
Figure~\ref{fig:expand} shows a partial unfolding of $(\evalo\ \lvar\expr\ \env\ \out)$.

%It is this unfolding of constraints that defines the
%search process.
As we unfold more nodes, branches of the constraint tree
constrain \lvar\expr\ to be more specific.
We call a partial specification of \lvar\expr\ as a ``candidate'' partial
program.
If at some point we find a fully specified \expr\ that satisfies all
relevant constraints, then \expr\ is a solution to the PBE problem.

In Figure~\ref{fig:model}, we show portions of the constraint tree
representing a PBE problem with two input/output pairs.
Each of the gray boxes corresponds to a separate disjunct in the constraint tree, representing a candidate.
Each disjunct is a conjunction of constraints, shown one on each line.
A candidate is viable only if the entire conjunction can be satisfied.
In the left column (a) certain
``obviously'' failing candidates like \texttt{(quote \lvar{M})} are omitted from consideration.
The right column (c) also shows the unfolding of
the selected disjunct for \texttt{(cons \lvar{D} \lvar{E})}, where \lvar{D} is replaced by its possible
values.

%In a standard implementation of miniKanren, the constraint tree is
%implicitly represented using suspended computations
%and program continuations.
%For this work, a language contributor built a transparent version of
%miniKanren to make the constraint tree explicit.

% TODO: fix this.
%\begin{table}[h]
% \centering
% \label{tab:constraints}
% \caption{Constraints Associated with Candidates for Examples
%          \{1 $\rightarrow$ (1 1 1), a $\rightarrow$ (a a a)\}}
% \begin{tabular}{p{0.1cm} p{0.1cm} l}
% \hline
%  \multicolumn{3}{l}{Candidate: ($\car$ $\lvar{s}_1$)} \\
%  & \multicolumn{2}{l}{\conj} \\
%  & $\arr$ & $\evalo$($\lvar{s}_1$ (1) ($\cons$ (1 1 1) $\lvar{t}_1$)) \\
%  & $\arr$ & $\evalo$($\lvar{s}_1$ (a) ($\cons$ (a a a) $\lvar{t}_1$)) \\
% \hline
%  \multicolumn{3}{l}{Candidate: ($\cdr$ $\lvar{s}_2$)} \\
%  & \multicolumn{2}{l}{\conj} \\
%  & $\arr$ & $\evalo$($\lvar{s}_2$ (1) ($\cons$ $\lvar{t}_2$ (1 1 1))) \\
%  & $\arr$ & $\evalo$($\lvar{s}_2$ (a) ($\cons$ $\lvar{t}_2$ (a a a))) \\
% \hline
%  \multicolumn{3}{l}{Candidate: ($\cons$ $\lvar{s}_3$ $\lvar{s}_4$)} \\
%  & \multicolumn{2}{l}{\conj} \\
%  & $\arr$ & $\evalo$($\lvar{s}_3$ (1) 1) \\
%  & $\arr$ & $\evalo$($\lvar{s}_4$ (1) (1 1)) \\
%  & $\arr$ & $\evalo$($\lvar{s}_3$ (a) a) \\
%  & $\arr$ & $\evalo$($\lvar{s}_4$ (a) (a a)) \\
% \hline
%  \multicolumn{3}{l}{Candidate: ($\const$ $\lvar{s}_5$)} \\
%  & \multicolumn{2}{l}{\conj} \\
%  & $\arr$ & $\lvar{s}_5$ $==$ (1 1 1) \\
%  & $\arr$ & $\lvar{s}_5$ $==$ (a a a) \\
% \hline
% \end{tabular}
%\end{table}

By default, miniKanren uses a biased interleaving search~\cite{byrd2017},
%The default search process used by miniKanren is biased interleaving,
alternating between disjuncts to unfold.
The alternation is ``biased'' towards disjuncts that have more of
their constraints already satisfied.
This search is \textit{complete}: if a solution exists,
it will eventually be found, time and memory permitting.
%For this work, we make one further transformation to the constraint
%tree representation, as in Table~\ref{tab:constraints}. We discard
%global tree structure, and associate subtrees with their corresponding
%candidate partial program $\p$. We will use the subtree of constraints
%associated with each candidate to score candidates.
%This way, we only concern
%ourselves with constraints associated with each candidate, thus being
%able to score candidates and think in terms of candidates.

%Some of the constraints in Table~\ref{tab:constraints},
%specifically those for the candidate program ($\const$ $\lvar{s}_5$),
%are obviously not satisfiable. In such cases, miniKanren will remove
%the candidate partial program and its associated constraints from
%consideration, so the machine learning guide will not need to score
%obviously bad candidates.

\subsection{Transparent constraint representation}

%\subsection{The miniKanren language}

%Make a chart here?

%The miniKanren language expresses programs in terms of logical constraints.  It provides values such as booleans, numbers, symbols, null, and pairs; logic variables for representing unknowns, which may be embedded in pairs; an equality constraint, asserting that two values are equal; forms for introducing new logic variables, forming conjunctions, and forming disjunctions; a way to define new relations, representing potentially-recursive constraints; and a way to query a system of constraints for logic variable assignments that satisfy it.

%Need to talk about how constraint trees relate to the relations.
%Start from calling relations constraints?

%In a standard implementation of miniKanren, the constraint tree is
%implicitly represented using suspended computations
%and program continuations.
%For this work, a language contributor built a transparent version of
%miniKanren to make the constraint tree explicit.

Typical implementations of miniKanren represent constraint trees as ``goals''~\cite{byrd2017} built from opaque, suspended computations.  These suspensions entangle both constraint simplification and the implicit search policy, making it difficult to inspect a constraint tree and experiment with alternative search policies.

%Fortunately, miniKanren implementations are fairly straightforward to
%build~\cite{hemann2013mukanren},
%so miniKanren has been re-implemented and extended by many.
One of our contributions is a miniKanren implementation that represents the constraint tree as a transparent data structure. It provides an interface for choosing the next disjunct to unfold, making it possible to define custom search policies driven by external agents.
Our implementation is available at
\href{https://github.com/xuexue/neuralkanren}{https://github.com/xuexue/neuralkanren}.

Like the standard miniKanren, this transparent version is implemented in Scheme.  To interface with an external agent, we have implemented a Python interface that can drive the miniKanren process via stdin/stdout. Users start by submitting a query, then alternate between receiving constraint tree updates and choosing the next disjunct to unfold.

%(TODO: refer to supplementary material)

\section{Neural Guided Constraint Logic Programming}

\begin{figure}[ht]
\begin{center}
\includegraphics[width=5.5in]{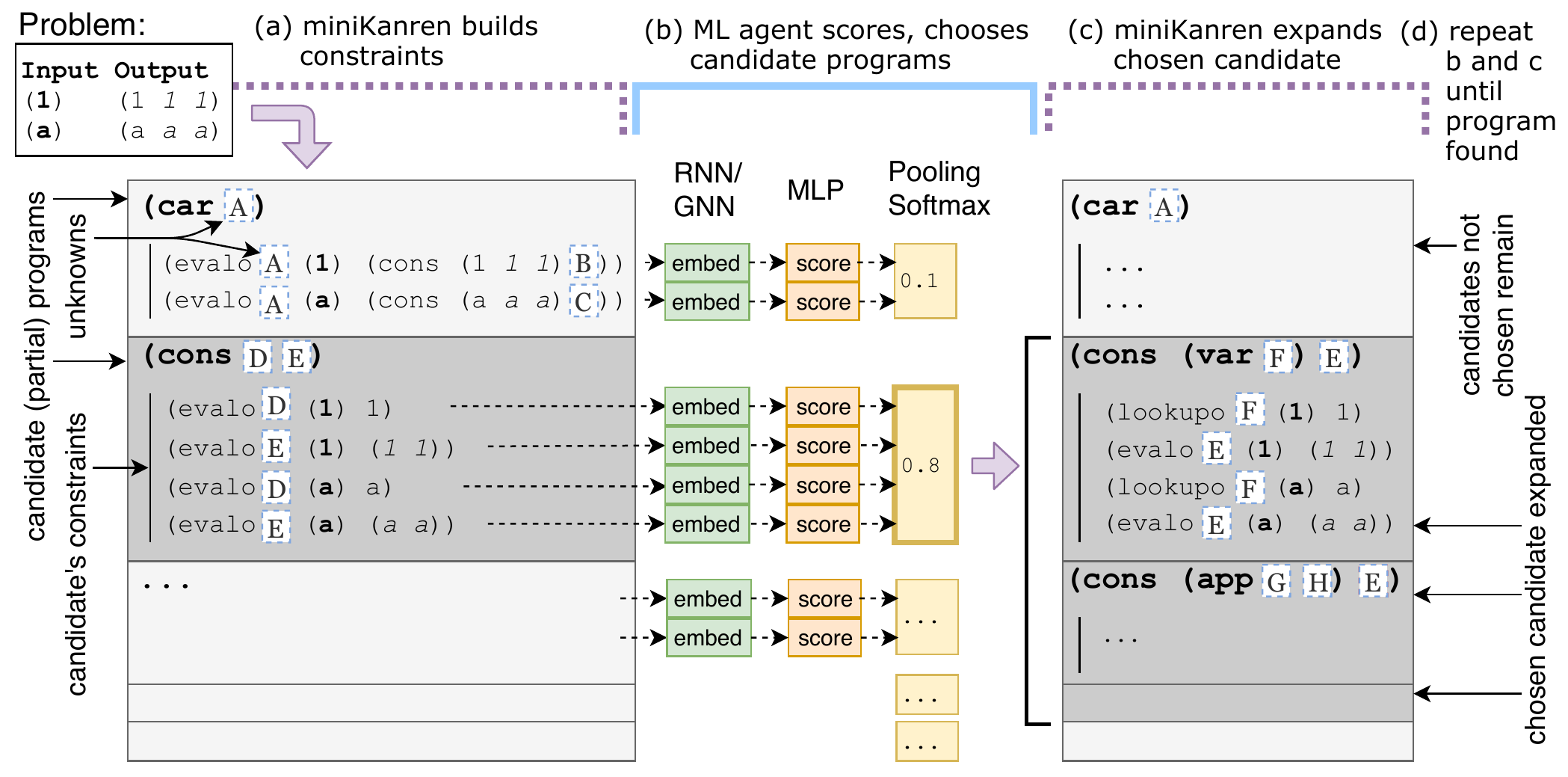}
\end{center}
\caption{Steps for synthesizing a program that repeats a symbol three times
using a subset of Lisp:
(a) miniKanren builds constraints representing the PBE problem;
    candidate programs contain unknowns, whose values are restricted by constraints;
(b) a neural agent operating on the constraints scores candidates;
    each constraint is embedded and scored separately, then pooled per candidate;
    scores determine which candidate to expand;
(c) miniKanren expands the chosen candidate \texttt{(cons \protect\lvar{D} \protect\lvar{E})}; possible completions of
    unknown \texttt{\protect\lvar{D}} are added to the set of candidates;
(d) this process continues until a fully-specified program (with no logic variables) is found.
    %a candidate with no unknowns satisfies all constraints imposed by the input/output examples.
    }
\label{fig:model}
\end{figure}

We present our neural guided synthesis approach summarized in Figure~\ref{fig:model}.
To begin, miniKanren represents the PBE problem in terms of a disjunction of 
candidate partial programs, and the constraints
that must be satisfied for the partial program to be consistent with the examples.
A machine learning agent makes discrete choices amongst the possible candidates.
The symbolic system then expands the chosen candidate, adding expansions of the
candidate to the list of partial programs. % Figure~\ref{fig:setup} also depicts the
%relationship between miniKanren and the ML Agent.

% Optional:
The machine learning model follows these steps:

\begin{enumerate}
\item \textbf{Embed} the constraints. Sections~\ref{sec:rnn} and~\ref{sec:gnn} discuss
    two methods for embedding constraints that
    trade off ease of training and accounting for logic variable identity.

\item \textbf{Score} each constraint. Each constraint embedding is scored independently,
    using a multi-layer perceptron (MLP).
\item \textbf{Pool} scores together. We pool constraint scores for each candidate.
    We pool hierarchically 
    using the structure of the constraint tree, max-pooling along a disjunction and 
    average-pooling along a conjunction.
    We find that using average-pooling instead of min-pooling helps gradient flow.
    In Figure~\ref{fig:model} there are no internal disjunctions.
    %(Conjunctions are more frequently seen in the constraint tree,
    %and we find that using max-pooling hinders effective gradient flow.)
\item \textbf{Choose} a candidate.  We use a softmax distribution over
    candidates during training and choose greedily during test.
\end{enumerate}

Intuitively, the pooled score for each candidate represents the plausibility of constraints associated with
a candidate partial program being satisfied. So in some sense we are learning a neural
constraint satisfaction system in order to solve synthesis problems.

\subsection{Recurrent Neural Network Model (RNN)} \label{sec:rnn}

One way to embed the constraints is using an RNN operating
on each constraint as a sequence.  We use an RNN with bi-directional LSTM units
\cite{hochreiter1997long} to score constraints, with each constraint
separately tokenized and embedded.
The tokenization process removes identifying information of logic variables,
and treats all logic variables as the same token. While logic variable
identity is important,  since each
constraint is embedded and scored separately, the logic variable identity
is lost.

We learn separate RNN weights for each relation
(\evalo, \lookupo, etc). The particular set of constraint
types differs depending on the target synthesis language.

%We softmax over the scores for each candidate program, then make a
%discrete choice to predict the optimal candidate to expand in the following step.

\subsection{Graph Neural Network Model (GNN)} \label{sec:gnn}

%The RNN model has the advantage of simplicity and performance. However,
In the RNN model, we lose considerable information by removing the identity of logic
variables. Two constraints associated with a logic variable
may independently be satisfiable, but may be obviously unsatisfiable
together. % Identifying information about logic variables should be
% important for more difficult synthesis problems.

To address this, we use a GNN model
that embeds all constraints simultaneously.
The use of graph or tree structure to represent programs~\cite{allamanis2018learning, chen2018tree}
and constraints~\cite{selsam2018learning} is not unprecedented.
An example graph structure is shown in Figure~\ref{fig:gnnconstr}.
Each constraint is represented as a tree, but since
logic variable leaf nodes may be shared by multiple constraints,
the constraint graph
is in general a Directed Acyclic Graph (DAG).
We do not include the
constraint tree structure (disjunctions and conjunctions) in the graph
structure since they are handled during pooling.

\begin{figure}[ht]
    \centering

\begin{tikzpicture}[sibling distance=20em,
  every node/.style = {shape=rectangle, rounded corners, draw, align=center, node distance=1.5cm, color=white, text=black}
  ]]
  \tikzstyle{evalo}=[text=black, color=orange]
  \tikzstyle{lvar}=[]
  \tikzstyle{etc}=[]
  \tikzstyle{cons}=[color=purple]

  \node [evalo] (e1) {$\evalo$};
  \node [lvar] (s1) [below left of = e1, xshift=-2mm] {\lvar{A}};
  \node [cons] (p1) [below of = e1, yshift=4.5mm] {\cons};
  \node [etc] (p1l) [below left of = p1, xshift=6mm, yshift=4mm] {1};
  \node [etc] (p1r) [below right of = p1, xshift=-6mm, yshift=4mm] {()};
  \node [cons] (c1) [below right of = e1, xshift=2mm] {$\cons$};

  \node [lvar] (t1) [below right of = c1, yshift=1mm] {\lvar{B}};

  \node [cons] (p111) [below of = c1, yshift=4mm] {$\cons$};
  \node [etc] (p111l) [below left of = p111, xshift=6mm, yshift=4mm] {1};
  \node [cons] (p111r) [below right of = p111, xshift=-6mm, yshift=4mm] {\cons};
  \node [etc] (p111rl) [below left of = p111r, xshift=6mm, yshift=4mm] {1};
  \node [cons] (p111rr) [below right of = p111r, xshift=-6mm, yshift=4mm] {\cons};
  \node [etc] (p111rrl) [below left of = p111rr, xshift=6mm, yshift=4mm] {1};
  \node [etc] (p111rrr) [below right of = p111rr, xshift=-6mm, yshift=4mm] {()};
  \node [evalo] (e2) [right of = e1, xshift=25mm] {$\evalo$};
  \node [cons] (a) [below of = e2, yshift=4.5mm] {\cons};
  \node [etc] (al) [below left of = a, xshift=6mm, yshift=4mm] {a};
  \node [etc] (ar) [below right of = a, xshift=-6mm, yshift=4mm] {()};

  \node [cons] (c2) [below right of = e2, xshift=2mm] {$\cons$};
  \node [cons] (aaa) [below of = c2, yshift=4mm] {\cons};
  \node [etc] (aaal) [below left of = aaa, xshift=6mm, yshift=4mm] {a};
  \node [cons] (aaar) [below right of = aaa, xshift=-6mm, yshift=4mm] {\cons};
  \node [etc] (aaarl) [below left of = aaar, xshift=6mm, yshift=4mm] {a};
  \node [cons] (aaarr) [below right of = aaar, xshift=-6mm, yshift=4mm] {\cons};
  \node [etc] (aaarrl) [below left of = aaarr, xshift=6mm, yshift=4mm] {a};
  \node [etc] (aaarrr) [below right of = aaarr, xshift=-6mm, yshift=4mm] {()};

  \node [lvar] (t2) [below right of = c2, yshift=1mm] {\lvar{C}};

  \path (e1) edge   (s1)
             edge   (p1)
             edge   (c1)
        (e2) edge   (s1)
             edge   (a)
             edge   (c2)
        (a) edge (al) edge (ar)
        (p1) edge (p1l) edge (p1r)
        (p111) edge (p111l) edge (p111r)
        (p111r) edge (p111rl) edge (p111rr)
        (p111rr) edge (p111rrl) edge (p111rrr)
        (c1)  edge   (p111)
              edge   (t1)
        (c2)  edge   (aaa)
              edge   (t2)
        (aaa) edge (aaal) edge (aaar)
        (aaar) edge (aaarl) edge (aaarr)
        (aaarr) edge (aaarrl) edge (aaarrr);

\end{tikzpicture}
    \caption{Graph representation of constraints \texttt{(evalo~\protect\lvar{A}~(1)~(cons~(1~1~1)~\protect\lvar{B}))} and \texttt{(evalo~\protect\lvar{A}~(a)~(cons~(a~a~a)~\protect\lvar{C}))}}
    \label{fig:gnnconstr}
\end{figure}

The specific type of GNN model we use is a Gated Graph Neural Network (GGNN) \cite{li2015gated}.
Each node has an initial embedding, which is refined through message passing
along the edges. The final root node embedding of each constraint 
is taken to be the embedding representation of the constraint.
Since the graph structure is a DAG,
we use a synchronous message schedule for message passing.

%We propagate messages first from
%the leaves of the tree upwards towards the root, then from the root
%of the tree down to the leaves. Each node updates its own embedding
%prior to sending messages further up/down the tree.

One difference between our algorithm and a typical GGNN is the use
of different node types. Each token in the constraint tree
(e.g. $\evalo$, $\cons$, logic variable)
has its own aggregation function and Gated Recurrent Unit weights.
Further, the edge types will also follow the node type of the
parent node.  Most node types will have asymmetric children,
so the edge type will also depend on the position of the child.

To summarize, the GNN model has the following steps:

\begin{enumerate}
    \item \textbf{Initialization} of each node, depending on the node type and label.
    The initial embeddings $e_{label}$ are learned parameters of the model.
    \item \textbf{Upward Pass}, which is ordered leaf-to-root,
    so that a node receives all
    messages from its children and updates its embedding before sending
    a message to its parents. Since a non-leaf node always has a fixed
    number of children, the merge function is parameterized
    as a multi-layer perceptron (MLP) with a fixed size input.
    \item \textbf{Downward Pass}, which is ordered root-to-leaf,
    so that a node receives all messages
    from its parents and updates its embedding before sending a message to
    its children. Nodes that are not logic variables will only have one parent,
    so no merge function is required. Constant embeddings are never updated.
    Logic variables can have multiple parents, so an average pooling is used
    as a merge function.
    \item \textbf{Repeat}. The number of upward/downward passes is a
    hyperparameter.
    We end on an upward pass so that logic variable updates
    are reflected in the root node embeddings.
\end{enumerate}

We extract the final embedding of the constraint root nodes for scoring,
pooling, and choosing.

\subsection{Training} \label{sec:training}

We note the similarity in the setup to a Reinforcement Learning problem.
The candidates can be thought of as possible \textit{actions},
the ML model as the \textit{policy}, and miniKanren
as the non-differentiable \textit{environment} which produces the \textit{states} or
constraints. However, during training we have access to the ground-truth
optimal action at each step, and therefore use a supervised cross-entropy loss.

We do use other techniques from the Reinforcement Learning literature.
We use curriculum learning, beginning with simpler training problems.
We generate training states by using the current model parameters
to make action choices at least some of the time. We use scheduled sampling
\cite{bengio2015scheduled} with a linear schedule,
to increase exploration and reduce teacher-forcing as training progresses.
We use prioritized experience replay \cite{schaul2015prioritized}
to reduce correlation in a minibatch, and re-sample more difficult states.
%storing explored states into a replay buffer and sampling mini-batches of
%states from this buffer.
%Experience replay is useful for reducing correlation in a minibatch,
%since nearby states in the same problem tend to be highly correlated.
%Priority sampling, re-sampling states with higher loss in a previous iteration
%wigh higher probability, has been shown to improve sample efficiency and
%improve the speed of training.
To prevent an exploring agent from becoming ``stuck'', we abort episodes
after 20 consecutive incorrect choices.
%when the agent chooses the wrong candidate 20 consecutive times.
For optimization we use RMSProp~\cite{Tieleman2012}, with weight decay
for regularization.

Importantly, we choose to expand two candidates per step during training,
instead of the single candidate as described earlier. % , allowing miniKanren to expand both candidates.
We find that expanding two candidates during training allows a better
balance of exploration / exploitation during training, leading to a more
robust model.
During test time, we resume expanding one candidate per step, and use a
greedy policy.

%\section{Negative Results}
%This section describes alternative training and inference
%techniques that we experimented with.
%
%\subsection{Leaf Timeout}
%
%Basic idea: If a leaf has too low of a probability in a previous step,
%remove it from the computation tree the next step.
%Helps with inference time.
%
%\subsection{De-Normalized Tree}
%
%Basic idea: Push disjunctions to the top of the tree. Work on each leaf
%separately. Do multiple rounds of up-down passes.
%
%\subsection{Hierarchical sampling}
%
%Basic idea: Hierarchical sampling
%
%\subsection{Reinforcement Learning}
%

\section{Experiments}

Following the programming languages community, we focus on tree manipulation
as a natural starting point towards expressive computation.
We use a small subset of Lisp as our target language.
This subset consists of $\cons$, $\car$,  $\cdr$, along
with several constants and function application. The full
grammar is shown in Figure~\ref{fig:lispdsl}.
%The Supplementary Material provides a definition of \evalo\ used in our experiments.

\begin{figure}[h]
\begin{verbatim}
datum (D)         ::= () | #t | #f | 0 | 1 | x | y | a | b | s | (D . D)
variable-name (V) ::= () | (s . V)
expression (E)    ::= (var V) | (app E E) | (lambda E) | (quote D)
                    | (cons E E) | (car E) | (cdr E) | (list E ...)
\end{verbatim}
\caption{Subset of Lisp used in this work}
\label{fig:lispdsl}
\end{figure}

We present two experiments. First, we test on programmatically
generated synthesis problems held out from training. We compare two miniKanren
search strategies that do not use a neural guide, three of our neural-guided models,
and RobustFill with a generous beam size.
Then, we test the generalizability of these approaches on three families of
synthesis problems. In this second set of experiments we additionally compare against
state-of-the-art systems $\lambda^2$, Escher, and Myth.
All test experiments are run on Intel i7-6700 3.40GHz CPU with 16GB RAM.

\subsection{Tree Manipulation Problems} \label{sec:lispprob}

We programmatically generate training data by querying $(\evalo\ \lvar\expr\ \lvar\env\ \lvar{\out})$ in miniKanren,
where the program, inputs, and outputs are all unknown. We put several other restrictions on
the inputs and outputs so that the examples are sufficiently expressive.
When input/output expressions contain constants,
we choose random constants to ensure variety. We use 500 generated problems for training,
each with 5 input/output examples.
In this section, we report results on 100 generated test problems.
We report results for several symbolic and neural guided models.
Sample generated problems are included in Appendix~\ref{sec:probs}.

% Some examples of generated problems are shown in Table~\ref{tab:lispeg}.

%\begin{table}
% \centering
% \label{tab:lispeg}
% \begin{tabular}{p{1cm} l l}
% \hline
%  \multicolumn{3}{l}{Program: \textsc{(lambda (car (car (var ()))))}} \\
%   & Input & Output \\
%   & \verb|((b . #t))| & \verb|b| \\
%   & \verb|((() . b) . a)| & \verb|()| \\
%   & \verb|((a . s) . 1)| & \verb|a| \\
%   & \verb|(((y . 1)) . 1)| & \verb|(y . 1)|\\
%   & \verb|((b))| & \verb|b| \\
% \hline
%  \multicolumn{3}{l}{Program: \textsc{(lambda (cons (car (var ())) (quote x)))}} \\
%   & Input & Output \\
%   & \verb|(a)| & \verb|(a . x)| \\
%   & \verb|(#t . s)| & \verb|(#t . x)| \\
%   & \verb|((1 . y) . y)| & \verb|((1 . y) . x)| \\
%   & \verb|((y 1 . s) . 1)| & \verb|((y 1 . s) . x)| \\
%   & \verb|(((x . x)) . y)| & \verb|(((x . x)) . x)| \\
% \hline
%  \multicolumn{3}{l}{Program: \textsc{(lambda (quote x))}} \\
%   & Input & Output \\
%   & \verb|y| & \verb|x| \\
%   & \verb|()| & \verb|x| \\
%   & \verb|#t| & \verb|x| \\
%   & \verb|a| & \verb|x| \\
%   & \verb|b| & \verb|x| \\
% \hline
% \end{tabular}
% \caption{Example auto-generated training problems using Lisp grammar.
%          The variables in a function are encoded using de Bruijn indices.
%          The symbol . denotes a pair.}
%\end{table}

We compare two variants of symbolic methods that use miniKanren.
The ``Naive'' model uses biased-interleaving search, as described
in \cite{byrd2012miniKanren}.  The ``+ Heuristic'' model uses additional
hand tuned heuristics described in \cite{byrd2017}.
The neural guided models include the RNN+Constraints guided search described
in Section~\ref{sec:rnn} and the GNN+Constraints guided search in Section~\ref{sec:gnn}.
The RNN model uses 2-layer bi-directional LSTMs with embedding size of 128.
The GNN model uses a single up/down/up pass with embedding size 64 and message
size 128. Increasing the number of passes did not yield improvements.
Further, we compare against a baseline RNN
model that does not take constraints as input:
instead, it computes embeddings of the input, output, and the candidate
partial program using an LSTM, then scores the concatenated embeddings
using a MLP. This
baseline model also uses 2-layer bi-directional LSTMs with embedding size of 128.
All models use a 2-layer neural network with ReLU activation as the scoring
function.
% None of the RNN models, including the baseline model, use attention.

Table~\ref{table:lisptestprob} reports the percentage of problems solved within 200 steps.
The maximum time the RNN-Guided search used was 11 minutes, so we allow the
symbolic models up to 30 minutes. The GNN-Guided search is significantly
more computationally expensive, and the RNN baseline model (without constraints)
is comparable to the RNN-Guided models (with constraints as inputs).
%Results are shown in Table~\ref{table:lisptestprob}.

\begin{table}[h]
\centering
\caption{Synthesis Results on Tree Manipulation Problems}
\label{table:lisptestprob}
\begin{tabular}[h]{l|ll}
\hline
Method                           & Percent Solved & Average Steps \\
\hline
Naive \cite{byrd2012miniKanren}  & 27\%           & N/A \\
+Heuristics (Barliman) \cite{byrd2017}      & 82\%           & N/A \\
\hline
RNN-Guided (No Constraints)      & 93\%           & 46.7 \\
GNN-Guided + Constraints         & 88\%           & 44.5 \\
RNN-Guided + Constraints         & 99\%           & \textbf{37.0} \\
\hline
RobustFill \cite{devlin2017robustfill} beam 1000+
                                 & \textbf{100\%} & N/A \\
\hline
\end{tabular}
\end{table}

All three neural guided models performed better than symbolic methods in
our tests, with the RNN+Constraints model solving all but one problem.
%The GNN-Guided + Constraints model was particularly time-consuming to train.
The RNN model without constraints also performed reasonably, but took
more steps on average than other models. RobustFill \cite{devlin2017robustfill}
Attention-C with large beam size solves one more problem
than RNN+Constraints on a flattened representation of these problems.
Exploration of beam size is in Appendix~\ref{sec:robustfill}.
% Figure~\ref{fig:steps} shows the number of
%actual vs optimal steps taken by the three models. The RNN+Constraints model
%is worse at smaller problems, but better overall. The RNN model without
%constraints typically takes fewer steps, but there are certain problems
%with 12 optimal steps where the model takes close to the cutoff of 200 steps.
We defer comparison with other symbolic systems because % outside of the miniKanren
% family because $\lambda^2$, Escher, and Myth all assume a strongly typed DSL, whereas
problems in this section involve dynamically-typed,
improper list construction. %Further, these symbolic systems do not

\subsection{Generalizability}

In this experiment, we explore generalizability. We use the same model weights as above
to synthesize three families of programs of varying complexity: \texttt{Repeat(N)} which
repeats a token $N$ times, \texttt{DropLast(N)} which drops the last element in an $N$ element
list, and \texttt{BringToFront(N)} which brings the last element to the front in an $N$
element list.  As a measure of how synthesis difficulty increases with $N$,
\texttt{Repeat(N)} takes $4+3N$ steps,
\texttt{DropLast(N)} takes $\frac{1}{2}N^2+\frac{5}{2}N+1$ steps, and
\texttt{BringToFront(N)} takes $\frac{1}{2}N^2+\frac{7}{2}N+4$ steps.
The largest training program takes optimally 22 steps to synthesize.
The number of optimal steps in synthesis correlates linearly with program size.

We compare against state-of-the-art systems $\lambda^2$, Escher, and Myth.
It is difficult to compare our models against other systems fairly, since these
symbolic systems use type information, which provides an advantage.
%For \texttt{Repeat(N)}, \texttt{DropLast(N)} and
%\texttt{BringToFront(N)}, typed systems should have an advantage.
Further, $\lambda^2$ assumes advanced language constructs like
\texttt{fold} that other methods do not. Escher is built as an active learner,
and requires an ``oracle'' to provide outputs for additional inputs. We
do not enable
this functionality of Escher, and limit the number of input/output examples to 5 for all methods.
We allow every method up to 30 minutes.
We also compare against RobustFill Attention-C with a beam size of 5000, the
largest beam size supported by our test hardware.
Our model is further restricted to 200 steps for
consistency with Section~\ref{sec:lispprob}.

Note that if given the full 30 minutes, the RNN+Constraints model is able to
synthesize \texttt{DropLast(7)} and \texttt{BringToFront(6)},
and the GNN+Constraints model is also able to synthesize \texttt{DropLast(7)}.
Myth solves \texttt{Repeat(N)} much faster than our model,
taking less than 15ms per problem, but fails on \texttt{DropLast} and \texttt{BringToFront}. Results are shown in Table~\ref{table:general}.

In summary, the RNN+Constraints and GNN+Constraints models both
solve problems much larger than those seen in training. The results
suggest that using constraints helps generalization: though RobustFill
performs best in Section~\ref{sec:lispprob}, it does not generalize
to larger problems out of distribution; though RNN+Constraints and
RNN-without-constraints
perform comparably in Section~\ref{sec:lispprob}, the former
shows better generalizability. This is consistent with the observation
that as program sizes grow, the corresponding constraints grow more
slowly.

\begin{table}[h]
\centering
\caption{Generalization Results: largest $N$ for which synthesis succeeded,
and failure modes (out of \textbf{time}, out of \textbf{memory},
requires \textbf{oracle}, other \textbf{error})}
\label{table:general}
\begin{tabular}[h]{l|lll}
\hline
Method
          & \small{Repeat(N)}
                       & \small{DropLast(N)}
                                       & \small{BringToFront(N)} \\
\hline
Naive \cite{byrd2012miniKanren}
          & 6 (time)   & 2 (time)      & - (time) \\
+Heuristics \cite{byrd2017}
          & 11 (time)  & 3 (time)      & - (time) \\
RNN-Guided + Constraints
          & \textbf{20+}      & \textbf{6} (time) & 5 (time) \\
GNN-Guided + Constraints
          & \textbf{20+}      & \textbf{6} (time) & \textbf{6} (time) \\
RNN-Guided (no constraints)
          & 9 (time)          & 3 (time)    & 2 (time)    \\
\hline
$\lambda^2$ \cite{feser2015synthesizing}
          & 4 (memory)    & 3 (error)        & 3 (error) \\
Escher \cite{albarghouthi2013recursive}
          & 10 (error)    & 1 (oracle)       & - (oracle) \\
Myth \cite{osera2015type}
          & \textbf{20+}
                          & - (error)        & - (error) \\
\hline
RobustFill \cite{devlin2017robustfill} beam 1000
          & 1 
                          & 1                & - (error) \\

RobustFill \cite{devlin2017robustfill} beam 5000
          & 3 
                          & 1                & - (error) \\
\hline
\end{tabular}
\end{table}

%\section{Recursive Problems in Lisp}

%\section{FlashFill}
%
%In this section, we use our approach to solve FlashFill problems. As the dataset
%used in \cite{gulwani2011flashfill}, \cite{devlin2017robustfill} and others are
%proprietary, we were unable to obtain the same dataset. We thus use a similar
%dataset used in the Syntax-Guided Synthesis (SyGuS) PBE String competition
%\cite{alur2017sygus}, and compare against the competitors.
%
%Currently, our system can represent about one third of the problems, because
%we haven't yet implemented disunification.
%
%\section{Visualizing Results}

\section{Conclusion}

% Should give a better vibe
% Our experiments were challenging (other systems had trouble) and substantial
% and a stepping stone towards general computation

%We presented a neural guided synthesis model where the neural guide takes as input the
%internal constraint encoding of the PBE problem used by miniKanren.
%Using a small subset of the Lisp language, we show promising
%results in the model's ability to generalize to larger problems.

%We have integrated a neural guided synthesis model directly with miniKanren's
%constraint representation of PBE problems, and demonstrated the potential of this approach.
%Though we only work within a small subset of Lisp,
%our experiments are substantial enough to challenge existing synthesis methods,
%and show that our model's ability to generalize to larger problems is encouraging.
%These results indicate our approach is a promising stepping stone towards more general computation.

%Our transparent implementation of miniKanren should prove useful to other neural guided synthesis researchers who are interested in such an integration.

%The ability to encode recursive problems is available in miniKanren, so
%learning to guide recursive program synthesis is left as future work.

We have built a neural guided synthesis model that works directly with miniKanren's
constraint representations, and a transparent implementation of miniKanren available at
\href{https://github.com/xuexue/neuralkanren}{https://github.com/xuexue/neuralkanren}.
We have demonstrated the success of our approach on challenging
tree manipulation and, more importantly, generalization tasks.
These results indicate that our approach is a promising stepping stone towards more general computation.

\clearpage
\subsubsection*{Acknowledgments}
Research reported in this publication was supported in part
by the Natural Sciences and Engineering Research Council of Canada,
and the National Center For Advancing Translational Sciences of
the National Institutes of Health under Award Number OT2TR002517.
R.L. was supported by Connaught International Scholarship.
The content is solely the responsibility of the authors and does not
necessarily represent the official views of the funding agencies.

\bibliography{nips_2018}
\clearpage

\begin{appendices}
\appendix
\setcounter{table}{0}
\renewcommand{\thetable}{A\arabic{table}}

\section{Relational Interpreter} \label{sec:reln}

We include below the code for the relational interpreter, written in miniKanren.
For readability by machine learning audience, our main paper renames the inputs to the
relational interpreter: \texttt{expr} or expression is called \expr\ or \textit{program} in the main paper,
\texttt{env} or environment is called \env\ or \textit{input}, and \texttt{value} is called \out\ or \textit{output}.

\begin{verbatim}
(define-relation (evalo expr env value)
  (conde              ;; conde creates a disjunction
    ((fresh (body)    ;; fresh creates new variables and a conjunction
       (== `(lambda ,body) expr)            ;; expr is a lambda definition
       (== `(closure ,body ,env) value)))
    ((== `(quote ,value) expr))             ;; expr is a literal constant
    ((fresh (a*)
       (== `(list . ,a*) expr)              ;; expr is a list construction
       (eval-listo a* env value)))
    ((fresh (index)
       (== `(var ,index) expr)              ;; expr is a variable
       (lookupo index env value)))
    ((fresh (rator rand arg env^ body)
       (== `(app ,rator ,rand) expr)        ;; expr is a function application
       (evalo rator env `(closure ,body ,env^))
       (evalo rand env arg)
       (evalo body `(,arg . ,env^) value)))
    ((fresh (a d va vd)
       (== `(cons ,a ,d) expr)              ;; expr is a cons operation
       (== `(,va . ,vd) value)
       (evalo a env va)
       (evalo d env vd)))
    ((fresh (c vd)
       (== `(car ,c) expr)                  ;; expr is a car operation
       (evalo c env `(,value . ,vd))))
    ((fresh (c va)
       (== `(cdr ,c) expr)                  ;; expr is a cdr operation
       (evalo c env `(,va . ,value))))))
\end{verbatim}

\clearpage

\section{Example Generated Problems} \label{sec:probs}

Some examples of automatically generated problems are shown in Table~\ref{tab:lispeg}.
Variables in a function body are encoded using de Bruijn indices,
so that \texttt{(var ())} is looking up the 0th (and only) variable.
The symbol . denotes a pair.

\begin{table}[h]
 \centering
 \caption{Sample auto-generated training problems}
 \label{tab:lispeg}
 \begin{tabular}{p{1cm} l l}
 \hline
  \multicolumn{3}{l}{Program: \textsc{(lambda (car (car (var ()))))}} \\
   & Input & Output \\
   & \verb|((b . #t))| & \verb|b| \\
   & \verb|((() . b) . a)| & \verb|()| \\
   & \verb|((a . s) . 1)| & \verb|a| \\
   & \verb|(((y . 1)) . 1)| & \verb|(y . 1)|\\
   & \verb|((b))| & \verb|b| \\
 \hline
  \multicolumn{3}{l}{Program: \textsc{(lambda (cons (car (var ())) (quote x)))}} \\
   & Input & Output \\
   & \verb|(a)| & \verb|(a . x)| \\
   & \verb|(#t . s)| & \verb|(#t . x)| \\
   & \verb|((1 . y) . y)| & \verb|((1 . y) . x)| \\
   & \verb|((y 1 . s) . 1)| & \verb|((y 1 . s) . x)| \\
   & \verb|(((x . x)) . y)| & \verb|(((x . x)) . x)| \\
 \hline
  \multicolumn{3}{l}{Program: \textsc{(lambda (quote x))}} \\
   & Input & Output \\
   & \verb|y| & \verb|x| \\
   & \verb|()| & \verb|x| \\
   & \verb|#t| & \verb|x| \\
   & \verb|a| & \verb|x| \\
   & \verb|b| & \verb|x| \\
 \hline
  \multicolumn{3}{l}{Program: \textsc{(lambda (cons (car (var ())) (car (car (cdr (var ()))))))}} \\
   & Input & Output \\
     & \verb|(y (y . b) . y)| & \verb|(y . y)| \\
     & \verb|(x (1 . 1))| & \verb|(x . 1)| \\
     & \verb|(x ((y . a) . x) . a)| & \verb|(x y . a)| \\
     & \verb|((#f . #t) (#f . a) . 1)| & \verb|((#f . #t) . #f)| \\
     & \verb|(a ((y #f . #f) . 1) . a)| & \verb|(a y #f . #f)| \\
 \hline
   \multicolumn{3}{l}{Program: \textsc{(lambda (car (cdr (car (car (cdr (cdr (cdr (var ())))))))))}} \\
   & Input & Output \\
     & \verb|(#f a () ((#f b . 1) . y) . #t)| & \verb|b| \\
     & \verb|(x #t y ((() (#t . a) . s)))| & \verb|(#t . a)| \\
     & \verb|(x b s ((#f (s 1 . b) . y)) . s)| & \verb|(s 1 . b)| \\
     & \verb|(b () #f ((b ((x . #t) . x))) . a)| & \verb|((x . #t) . x)| \\
     & \verb|(1 #t a ((s (1 #t s . a) . x) . #t) . #t)| & \verb|(1 #t s . a)| \\
 \hline

\end{tabular}
\end{table}

\clearpage

\section{Problems where Neural Guided Synthesis Fails} \label{sec:fails}

Table~\ref{tab:fail} lists problems on which the
methods failed. The single problem that RNN + Constraints failed to solve
is a fairly complex problem. The problems that the GNN + Constraints
failed to solve all include a complex list accessor portion. This
actually makes sense: it is conceivable for multi-layer RNNs to be
better at this kind of problem compared to a single-layer GNN.
The RNN without constraints also fails at complex list accessor
problems.

\begin{table}[h]
\centering
\caption{Problems where Neural Guided Synthesis Fails}
\label{tab:fail}
\begin{tabular}[h]{l|l}
\hline
Method & Problem \\
\hline
RNN + Constraints & (lambda (cons (cons (var ()) (var ())) (cons (var ()) (car (cdr (var ())))))) \\
\hline
GNN + Constraints
& (lambda (car (car (car (car (cdr (cdr (car (var ()))))))))) \\
& (lambda (car (car (car (cdr (car (cdr (car (var ()))))))))) \\
& (lambda (car (car (car (cdr (cdr (cdr (car (var ()))))))))) \\
& (lambda (car (car (cdr (car (car (var ()))))))) \\
& (lambda (car (car (cdr (car (cdr (cdr (car (var ()))))))))) \\
& (lambda (car (car (cdr (cdr (cdr (cdr (car (var ()))))))))) \\
& (lambda (car (cdr (car (car (cdr (var ()))))))) \\
& (lambda (car (cdr (car (cdr (cdr (car (var ())))))))) \\
& (lambda (car (cdr (cdr (car (car (cdr (var ())))))))) \\
& (lambda (car (cdr (cdr (cdr (cdr (car (car (var ()))))))))) \\
& (lambda (car (cdr (cdr (cdr (cdr (car (cdr (var ()))))))))) \\
& (lambda (cdr (cdr (car (car (var ())))))) \\
\hline
RNN (No Constraints)
& (lambda (cons (car (var ())) (cons (var ()) (cdr (car (var ())))))) \\
& (lambda (cdr (car (car (cdr (car (cdr (var ())))))))) \\
& (lambda (cdr (car (cdr (car (car (car (var ())))))))) \\
& (lambda (cdr (car (car (car (car (car (var ())))))))) \\
& (lambda (cdr (car (car (cdr (car (cdr (var ())))))))) \\
& (lambda (cdr (car (cdr (car (car (car (var ())))))))) \\
& (lambda (cdr (car (car (car (car (car (var ())))))))) \\
\hline
\end{tabular}
\end{table}

\clearpage

\section{RobustFill Results for Various Beam Sizes} \label{sec:robustfill}

To compare against RobustFill, we use a flattened representation of the problems
shown in Section~\ref{sec:probs}, and use the Attention-C model with various beam
sizes. For a beam size $k$, if any of the top-$k$ generated programs are correct,
we consider the synthesis a success. We report several figures in Table~\ref{table:robustfill}:
column (a) shows the percent of test problems held out from training that
were successfully solved (Table 2 in our paper), and column (b) shows the largest $N$ for a 
family of synthesis problems for which synthesis succeeds (Table 3 in our paper).

\begin{table}[h]
\centering
\caption{RobustFill Results}
\label{table:robustfill}
\begin{tabular}[h]{l|l|llllll}
\hline
Model &  (a) Test & \multicolumn{3}{c}{(b) Generalization} \\

          & \small{\% Solved}
                     & \small{Repeat(N)}
                                     & \small{DropLast(N)}
                                                        & \small{BringToFront(N)} \\
\hline
RobustFill, Beam Size 1     &  56\% & 0 & 0 & 0 \\
RobustFill, Beam Size 10    &  94\% & 0 & 0 & 0 \\
RobustFill, Beam Size 100   &  99\% & 1 & 0 & 0 \\
RobustFill, Beam Size 1000  & 100\% & 1 & 1 & 0 \\
RobustFill, Beam Size 5000  & 100\% & 3 & 1 & 0 \\
\hline
RNN-Guided + Constraints (Ours)  & 99\% & 20+ & 6 & 5 \\
\hline
\end{tabular}
\end{table}

\end{appendices}

\end{document}